\documentclass[letterpaper]{article} 
\usepackage[]{aaai2026}  
\usepackage{times}  
\usepackage{helvet}  
\usepackage{courier}  
\usepackage[hyphens]{url}  
\usepackage{graphicx} 
\usepackage{natbib}  
\usepackage{caption} 
\usepackage{algorithm}
\usepackage{algorithmic}
\usepackage{newfloat}
\usepackage{listings}
\DeclareCaptionStyle{ruled}{labelfont=normalfont,labelsep=colon,strut=off} 
\floatstyle{ruled}
\newfloat{listing}{tb}{lst}{}
\floatname{listing}{Listing}
\usepackage{acronym}
\acrodef{dlgn}[DLGN]{Differentiable Logic Gate Networks}
\acrodef{gnn}[GNN]{Graph Neural Networks}
\acrodef{mlp}[MLP]{MultiLayer Perceptrons} %
\acrodef{dlc}[DLC]{Differentiable Logic Circuit}

\usepackage{booktabs}

\usepackage{amsmath}

\usepackage{multirow}
\usepackage{xcolor}
\usepackage{tikz}
\usepackage{pgfplots}
\pgfplotsset{compat=1.18} 

\usepackage{xcolor}

\usepackage{cleveref}

\title{GIC-DLC: Differentiable Logic Circuits \\ for Hardware-Friendly Grayscale Image Compression}
\author{
    Till Aczel,
    David F. Jenny,
    Simon Bührer,
    Andreas Plesner,
    Antonio Di Maio,
    Roger Wattenhofer \\
    ETH Zurich \\
}

\begin{document}

\maketitle

\begin{abstract}

Neural image codecs achieve higher compression ratios than traditional hand-crafted methods such as PNG or JPEG-XL, but often incur substantial computational overhead, limiting their deployment on energy-constrained devices such as smartphones, cameras, and drones.  
We propose \textbf{Grayscale Image Compression with Differentiable Logic Circuits (GIC-DLC)}, a hardware-aware codec where we train lookup tables to combine the flexibility of neural networks with the efficiency of Boolean operations.
Experiments on grayscale benchmark datasets show that GIC-DLC outperforms traditional codecs in compression efficiency while allowing substantial reductions in energy consumption and latency.
These results demonstrate that learned compression can be hardware-friendly, offering a promising direction for low-power image compression on edge devices.

\end{abstract}


\section{Introduction}

Learned image compression has demonstrated superior compression ratios compared to traditional hand-crafted codecs
by optimizing the compression objective in an end-to-end manner on datasets of images, predicting per-pixel distributions, and leveraging data-driven structure to minimize storage requirements \citep{balle2016end}.  
However, high compression performance often comes at a cost.  
Large neural networks perform many unnecessary operations, resulting in increased latency and energy consumption.
This is particularly problematic for edge devices such as smartphones, cameras, or drones, which have limited battery, weaker processors, and stricter thermal constraints than desktop or cloud hardware.

Traditional codecs such as PNG \citep{png}, WebP \citep{webp}, or JPEG-XL \citep{alakuijala2019jpegxl} remain fast and energy-efficient.  
They achieve this through hand-designed algorithms optimized for simplicity and speed, often supported by dedicated hardware units that accelerate key operations.
However, these methods lack the adaptability of learned models, which limits their compression performance compared to modern learned approaches.

This trade-off motivates the search for methods that have the efficiency of traditional codecs with the compression ratio of learned codecs.
To this end, we focus on trainable architectures operating closer to hardware, enabling direct deployment on edge devices for low-latency, low-power operation.
Differentiable Logic Circuits (DLCs) offer a natural solution.
By training lookup tables or logic circuits, DLCs combine the flexibility of neural networks with hardware-friendly computation that is fast yet remains energy-efficient.
They can be optimized end-to-end using gradient-based methods for task-specific objectives, yet remain directly deployable on devices such as FPGAs.

Leveraging DLCs, we propose \textbf{Grayscale Image Compression with Differentiable Logic Circuits (GIC-DLC)}, a neural codec that predicts per-pixel probability distributions in a hardware-aware manner.
Replacing conventional floating-point networks with trainable lookup tables and logic gates, GIC-DLC achieves high compression efficiency while maintaining low-latency, low-power inference, making it particularly suitable for energy-constrained devices.

We show that GIC-DLC achieves higher compression ratio, processing speed, and energy efficiency than traditional codecs on EMNIST \cite{cohen2017emnist} images.
These results demonstrate that differentiable logic offers a promising path toward trainable, hardware-friendly image compression, particularly for energy-constrained devices, and lay the foundation for future extensions to more complex natural images.

\begin{figure*}[t]
    \centering
    \includegraphics[width=\textwidth]{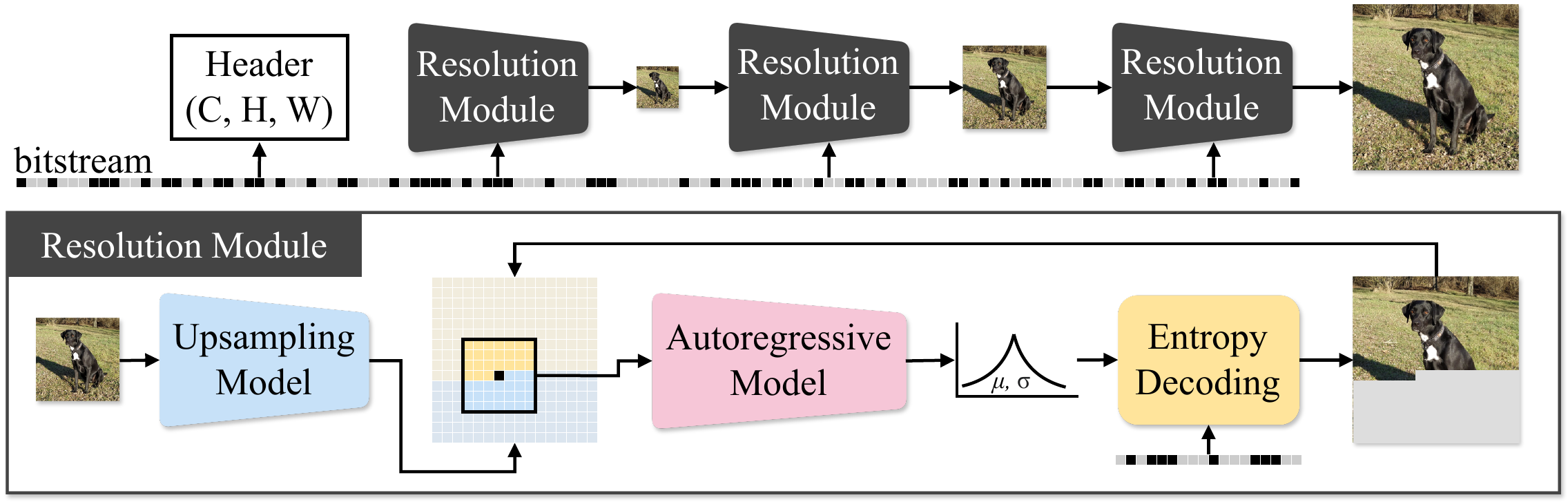}
    \caption{
    Overview of the decoding architecture.  
    Top: the image is upsampled $L=2$ times, with each higher-resolution level decoded conditioned on the lower-resolution one. 
    Bottom: at each resolution, an upsampling model (UPS) and an autoregressive model (ARM), both implemented as NeuraLUT networks, produce predictions used for entropy decoding.  
    The UPS model  outputs point estimates for the upsampled pixels (conditioned on the lower-resolution reconstruction), which serve as priors for pixels that are not yet decoded.  
    The ARM predicts per-pixel, Laplace parameters conditioned on the local, already-decoded context; when a pixel is not yet decoded, the ARM uses the UPS point estimate as its prior.  
    Decoding proceeds from the coarsest to the finest resolution; encoding follows the reverse order.
    }
    \label{fig: system}
\end{figure*}

\section{Related Work}

Classical image compression algorithms rely on hand-crafted pipelines that exploit simple statistical properties of natural images.  
Early formats such as JPEG~\citep{wallace1992jpeg} and PNG~\citep{png} achieve compression through linear transforms, quantization, and entropy coding.
Lightweight formats such as QOI~\citep{qoi2021} omit entropy coding and use simple predictors, enabling fast, low-cost lossless compression.
Modern successors such as WebP~\citep{webp} and JPEG-XL~\citep{alakuijala2019jpegxl} introduce advanced predictors and context-adaptive entropy models, improving compression ratios while maintaining low latency and energy efficiency.  
Many devices have been designed to accelerate traditional codecs such as JPEG through dedicated hardware.  
However, hand-crafted codecs rely on heuristic algorithms and, without data-driven learning, can only approach optimal compression.  

Learned image compression replaces hand-crafted transforms with trainable neural networks optimized end-to-end for rate–distortion performance \citep{balle2016end, mentzer2020high, he2022po}.  
While most research focuses on lossy compression, neural approaches have also been applied to lossless or near-lossless settings by modeling the full pixel distribution and compressing it using entropy coding \citep{mentzer_practical_2020, super-resolution-cao}.  
Autoregressive models are commonly used to predict the distribution of image data.
Raster-based architectures, such as PixelCNN \citep{van2016conditional}, sequentially predict pixels.
Hierarchical codecs, like SReC \citep{super-resolution-cao}, capture dependencies across latent levels in a structured residual framework.

Overfitted codecs, such as Cool-Chic~\citep{ladune2023cool} and C3~\citep{kim2024c3}, are designed to lower the decoding cost learned codecs while preserving their high compression performance.  
They encode by overfitting an entire network to the image and decode via inference.  
FNLIC~\citep{zhang2025fitted} extends this approach as a lossless codec.  
Despite fast decoding, encoding remains computationally expensive, requiring substantial per-image optimization and limiting the methods' use in scenarios demanding efficient encoding.  
Although recent work has reduced encoding time from over ten minutes on a GPU to a few seconds on a CPU~\citep{blard2024overfitted, borrell2025hypercool}, these methods still achieve worse compression ratio than traditional codecs.
A key limitation is that these methods rely on neural networks using floating-point computation, which are energy-intensive and involve redundant operations, restricting their applicability in energy-constrained environments.

Differentiable Logic Circuits (DLCs) provide a framework to learn from data while remaining hardware- and energy-efficient.  
They follow MLP-like structural principles but replace continuous arithmetic with differentiable logic or lookup operations.  
During training, a continuous ``soft'' approximation is used, which is discretized after convergence to produce a deployable ``hard'' circuit.  
Multiple frameworks have been proposed demonstrating different design choices and capabilities ~\citep{NEURIPS2022_0d3496dd, ruttgers_light_2025, bacellar_differentiable_2025}.  
For instance, NeuraLUT~\citep{andronic_neuralut_2024} uses a neural network to implement the soft approximation of the lookup table, improving scalability and expressiveness.  
DLCs have been applied to tasks such as image classification \citep{NEURIPS2024_db988b08} and natural language processing \citep{buhrer_recurrent_2025}, illustrating their versatility.  
Despite remaining challenges in training and scalability \citep{yousefi_mind_2025}, DLCs have been shown to achieve lower latency and energy efficiency than MLPs and CNNs while maintaining comparable accuracy \citep{NEURIPS2022_0d3496dd, bacellar_differentiable_2025}, highlighting their potential for hardware-oriented neural compression.

\section{Method}

Following other learned compression methods, we perform lossless image compression by predicting per-pixel probability distributions and encoding them into a bitstream with Asymmetric Numeral Systems (ANS) \citep{duda2013asymmetric}.  
This ensures likely pixels use fewer bits and unlikely ones more, producing a compact, lossless representation.  
Encoding and decoding are exact inverses using ANS.

\subsection{Hierarchical Compression Framework}

To capture image structure across scales, each image is decomposed into $L+1$ resolution levels for hierarchical encoding and decoding (see \Cref{fig: system}), similar to the approach used in SReC \citep{super-resolution-cao}.
Given an image $x \in \{0, \ldots, 255\}^{H \times W \times C}$, with height $H$, width $W$, and $C$ channels, we iteratively apply $2\times2$ average pooling to halve the spatial resolution at each level:
\[
x^{(\ell)} = \texttt{avgpool2}(x^{(\ell-1)}), \quad \ell = 1, \dots, L,
\]  
with $x^{(0)} = x$ being the original image.  
After pooling, pixel values at each level are rounded to integers.  

Decoding begins at the coarsest resolution, which is first reconstructed to provide an initial image approximation.  
Finer levels are then decoded in two stages: upsampling followed by autoregressive refinement, capturing both global and local dependencies.  
In contrast, SReC performs these two operations in a single stage.


\subsubsection{Upsampling Model (UPS)}

The UPS predicts the upsampled image $\tilde{x}^{(i)}$ independently of the bitstream.  
For each pixel in the lower-resolution image $x^{(i+1)}$, it simultaneously predicts a $2\times2$ block of pixels, conditioned only on a local neighborhood of size $K \times K$.  
This design efficiently propagates coarse structural information to higher resolutions.  
The network is implemented as a Differentiable Logic Circuit (DLC), and it is trained using a mean squared error loss.

\subsubsection{Autoregressive Model (ARM)}

The ARM refines the upsampled image by predicting per-pixel Laplace distribution parameters $\mu$ and $\sigma$.  
For pixels that have already been decoded, the ARM conditions on the local $K \times K$ context within the current resolution level.  
For pixels not yet decoded, it relies on the upsampled predictions $\tilde{x}^{(i)}$ as a prior.  
The ARM predicts individual Laplace parameters for all $C$ channels in parallel, allowing efficient multi-channel modeling.
At the coarsest level ($i=L$), no upsampling is required, and decoding starts from an empty initialization.  
The ARM is also implemented as a DLC network.

Following \cite{ladune2023cool}, the ARM works in a similar way and predicts per-pixel Laplace parameters $(\mu, \sigma)$.  
These continuous distributions are converted into probabilities over integer pixel values by integrating the Laplace density over each pixel’s quantization interval (e.g., from $x-0.5$ to $x+0.5$).  
The expected number of bits required to encode a pixel is then estimated as the negative log-probability of the true pixel value under this discrete distribution.  
This provides a differentiable proxy for the actual bitrate and serves as the training objective to improve compression efficiency.


\subsection{Differentiable Logic Circuits (DLCs)}

DLCs replace dense matrix multiplications with lightweight, trainable lookup tables, reducing computational overhead compared to standard MLPs.  
Inputs are binarized using thermometer encoding with 255 levels: for a pixel value $v \in [0,255]$, each level generates a binary indicator specifying whether $v$ exceeds the corresponding threshold, allowing DLCs to process continuous intensities via lookup tables.

For the learnable nodes, we use NeuraLUT \cite{andronic_neuralut_2024}, which replaces the static lookup table with a small neural network during training.  
The last layer of this network uses a sigmoid to constrain outputs to $[0,1]$.
To help the network learn binary outputs, we add logistic noise before the sigmoid, controlled by a temperature parameter $\tau_{\text{node}}$, which is annealed during training. 
When converting the network into a lookup table, all $2^6 = 64$ input combinations (for 6 binary inputs) are tested and stored, so inference reduces to a simple lookup.  

Connections between inputs and nodes are learned via a softmax over the learnable connection weights, with a temperature parameter $\tau_{\text{connections}}$ annealed during training to encourage harder selections.  
For evaluation, the final connections are fixed using an argmax over the learned weights.

The network's final output is the average of the last layer binary outputs, producing a float in $[0,1]$.  
For $\mu$, this is scaled to $[0,255]$, while $\sigma$ is transformed via a log inverse-sigmoid to expand its range.




\section{Experiment Setup}

The compression model is trained on the EMNIST ByClass dataset \citep{cohen2017emnist}, which contains grayscale images of digits as well as lowercase and uppercase letters.  
Evaluation is performed on the EMNIST test set and additionally on KMNIST \citep{clanuwat2018kmnist} and Fashion-MNIST (FMNIST) \citep{xiao2017fashionmnist} to assess generalization beyond the training distribution.  
Both the ARM and UPS networks consist of two layers with 1,024 lookup tables each.  
Training is carried out on 128,000 samples.  
Further details on the dataset and hyperparameters are provided in Appendix~\ref{appendix:experiment_setup}.

\section{Results Analysis}

\setlength{\tabcolsep}{5pt} 
\begin{table}[t]
    \small
    \centering
    \caption{Actual coded bits per pixel for different datasets. Our DLC-based method (GIC-DLC) outperforms the best traditional method (JPEG-XL) on the training distribution (EMNIST) while remaining fast and energy-efficient. IC-MLP is shown in gray for reference; the MLP-based learned image compression method achieves better compression but is orders of magnitude slower.}
    \begin{tabular}{lccc|cc}
    \toprule
    
         & \multicolumn{3}{c|}{in-distribution (EMNIST)} &\multicolumn{2}{c}{out-of-distribution} \\ 
         & all & letters & digits & KMNIST & FMNIST \\ 
    \midrule 
         QOI & 6.14 & 6.02 & 6.20 & 7.56 & 8.50 \\
         PNG & 4.18 & 4.13 & 4.16 & 4.45 & 5.18  \\
         WebP & 3.34 & 3.31 & 3.34 & \textbf{3.56} & 4.62\\
         JPEG-XL & 3.20 & 3.18 & 3.23 & 3.66 & \textbf{4.29} \\ 
        GIC-DLC & \textbf{2.74} & \textbf{2.78} & \textbf{2.71} & 4.16 & 6.27\\
    \midrule
        IC-MLP & 2.34 & 2.34 & 2.30 & 3.67 & 6.04 \\ 
    \bottomrule
    \end{tabular}
    \label{tab: main}
\end{table}
\setlength{\tabcolsep}{6pt} 

\Cref{tab: main} shows coded bits per pixel for different methods across datasets.  
Our method, GIC-DLC, outperforms traditional codecs on EMNIST, which matches the training distribution, surpassing even the best traditional method, JPEG-XL.  
The model is particularly effective on simpler digits, while letters are slightly harder to compress.  
On out-of-distribution data, the performance advantage decreases.  
On KMNIST, containing Korean characters, the model generalizes reasonably well, while on FMNIST, with more complex grayscale items, the bpp increases.  
This reflects the relatively narrow training distribution, suggesting that expanding the training set to cover more diverse images is a promising direction for future research.

\begin{figure}[t]
    \centering
    \hspace{-0.05\linewidth}
    \begin{minipage}{0.6\linewidth}
        \centering
        \begin{tikzpicture}
        \begin{axis}[
            ybar,
            bar width=8pt,
            width=\linewidth,
            height=4cm,
            ymin=0,
            ylabel={bpp-theoretical},
            ylabel style={font=\scriptsize, yshift=-4pt},
            symbolic x coords={level 0, level 1, level 2},
            xtick=data,
            enlarge x limits=0.3,
            xticklabel style={font=\scriptsize, yshift=4pt}, 
            yticklabel style={font=\scriptsize, xshift=2pt}, 
            xtick style={draw=none}, 
            legend style={at={(0.5,-0.2)}, anchor=north, legend columns=-1, font=\scriptsize}, 
        ]
        \addplot[fill=orange!80!black, draw=none] coordinates {(level 0,1.34) (level 1,0.58) (level 2,0.30)};
        \addplot[fill=blue!70!black, draw=none] coordinates {(level 0,1.66) (level 1,0.63) (level 2,0.30)};
        \legend{IC-MLP, GIC-DLC}
        \end{axis}
        \end{tikzpicture}
    \end{minipage}
    \hspace{-0.05\linewidth}
    \begin{minipage}{0.49\linewidth}
        \centering
        \begin{tikzpicture}
        \begin{axis}[
            ybar,
            bar width=8pt,
            width=\linewidth,
            height=4cm,
            ymin=0,
            ylabel={RMSE},
            ylabel style={font=\scriptsize, yshift=-4pt},
            symbolic x coords={level 1, level 2},
            xtick=data,
            enlarge x limits=0.6,
            xticklabel style={font=\scriptsize, yshift=4pt}, 
            yticklabel style={font=\scriptsize, xshift=2pt}, 
            xtick style={draw=none}, 
            legend style={at={(0.5,-0.2)}, anchor=north, legend columns=-1, font=\scriptsize}, 
        ]
        \addplot[fill=yellow!80!black, draw=none] coordinates {(level 1,11.90) (level 2,6.34)};
        \addplot[fill=blue!70!black, draw=none] coordinates {(level 1,2.65) (level 2,2.04)};
        \legend{Bicubic, GIC-DLC}
        \end{axis}
        \end{tikzpicture}
    \end{minipage}
    \caption{Comparison of theoretical bits per pixel (bpp-theoretical) across resolution levels (left) and root mean squared error (RMSE) of upsampling (right) on the EMNIST test set. The left plot shows that most bits are concentrated in the highest resolution, while the right plot highlights the superior accuracy of the learned upsampling.}
    \label{fig: level-upsampling-bar}
\end{figure}
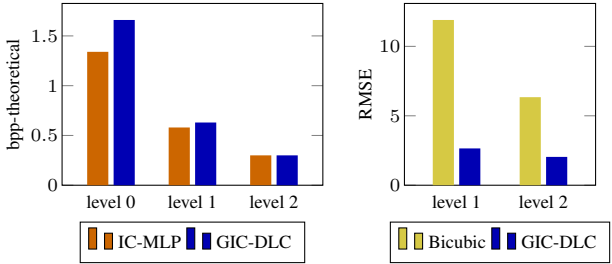


The multi-resolution design concentrates most information in the finest level, with coarser levels providing global context (\Cref{fig: level-upsampling-bar} left).  
Each finer level has four times more pixels, so most bits are naturally allocated to the highest resolution.  
This structure allows the ARM and UPS to efficiently model both local and global dependencies, improving compression without increasing computational cost.


Learned upsampling is crucial: the UPS network reduces reconstruction error compared to bicubic interpolation (\Cref{fig: level-upsampling-bar} right), capturing subtle structural details that standard upsampling misses.  
This improvement in prediction accuracy translates into lower bpp.


Overall, these results demonstrate that Differentiable Logic Circuits (DLCs) can be effectively applied to image compression.  
They achieve strong compression performance comparable to larger learned models (\Cref{tab: main}), while remaining fast and energy-efficient (\Cref{tab: energy_latency}).  
This addresses a significant limitation of traditional learned codecs, which often require substantial computation and are orders of magnitude slower.  
By combining upsampling with autoregressive refinement, DLC-based compression provides a practical solution for real-time and resource-constrained applications.

\section{Discussion}

\subsection{Energy Usage and Latency}
\label{subsec:energy_latency}

We estimate the energy usage and latency of PNG and our proposed method.  
\Cref{tab: energy_latency} summarizes the \emph{estimated} energy consumption and latency per pixel.   
Even under conservative assumptions, compared to PNG, our method is roughly 100 times more energy-efficient for encoding and 10 times for decoding compared.
Both encoding and decoding remain below 5ns per pixel, making them competitive in latency as well.  
These values are derived from literature-based estimates and analytical extrapolations, as detailed in Appendix~\ref{appendix:energy_estimation}, rather than from direct hardware measurements.  
Thus, they should be interpreted as indicative of expected efficiency rather than empirical measurements.

\begin{table}[t]
    \centering
    \caption{Estimated energy consumption and latency per pixel for PNG and our method.  
    Our approach drastically reduces both energy and latency, achieving near-instant encoding and decoding with minimal power, while PNG is orders of magnitude more expensive.}
    \label{tab: energy_latency}
    \begin{tabular}{lcc|cc}
        \toprule
        Method & \multicolumn{2}{c|}{Energy [nJ/pixel]} & \multicolumn{2}{c}{Latency [ns/pixel]} \\
        \cmidrule(lr){2-3} \cmidrule(lr){4-5}
               & Encode & Decode & Encode & Decode \\
        \midrule
        PNG & 322.58 & 39.19 & 44.99 & 5.17 \\
        Our Method & $\approx$4 & $\approx$4 & $<$5 & $<$5 \\
        \bottomrule
    \end{tabular}
\end{table}

\subsection{Potential Applications}

The high cost of neural codecs has limited their industrial adoption, despite superior compression performance.  
GIC-DLC addresses this by providing a hardware-aware, low-power alternative, making learned compression practical for energy- and latency-constrained applications.

Edge devices, such as smartphones, tablets, cameras, and drones, frequently perform image and video compression tasks.  
While decoding latency must meet real-time requirements, energy consumption and efficiency directly impact battery life, thermal performance, and device usability.  
By replacing floating-point networks with differentiable logic circuits, GIC-DLC reduces computation, enabling high compression ratios with low-latency, low-power operation.

This efficiency makes GIC-DLC particularly suitable for edge devices.  
Its low computational footprint also enables integration of learned compression in large-scale industrial systems without the prohibitive energy and hardware costs of traditional neural codecs, providing a practical path to high-performance, hardware-friendly image compression.

\subsection{Future Work }

While our method has been demonstrated on grayscale datasets, it remains to be tested on RGB images with natural image distributions to evaluate its performance on more complex, real-world data.  
Additionally, compiling the network and deploying it on an FPGA board will be essential to measure practical efficiency, including throughput, latency, and energy consumption in a hardware setting.

\section{Conclusion}

We have presented a compression framework that leverages multi-resolution priors, staged upsampling, and autoregressive refinement, implemented with Differentiable Logic Circuits (DLCs).  
Our approach achieves high compression performance on small, grayscale image datasets while remaining fast and energy-efficient, highlighting the potential of lightweight, trainable logic for learned compression.  

Although this study focuses on relatively simple toy datasets, the observed benefits of DLCs in balancing performance and efficiency provide strong motivation to explore their application to larger and more complex image distributions.  
This work thus serves as a stepping stone toward practical, fast, and low-energy learned compression methods for a wider range of images.


\newpage
\bibliography{references_manual}

\appendix

\section{Experiment Setup Detailed}
\label{appendix:experiment_setup}

\subsection{Datasets}

The compression model is trained on the training split of EMNIST ByClass, which includes grayscale images of digits as well as lowercase and uppercase letters.  
Evaluation is performed on the EMNIST test set and on two additional datasets: KMNIST and Fashion-MNIST.  

KMNIST contains grayscale images of handwritten Korean characters.  
Although it shares the same structure and resolution as EMNIST, the symbol set differs, making it an out-of-distribution dataset that is still relatively close in nature.  
In contrast, Fashion-MNIST consists of grayscale images of clothing items.  
These images exhibit much higher visual complexity, with curved and textured regions rather than simple line strokes, representing a significantly different data distribution.

\subsection{Baselines}

We compare our approach against several traditional lossless compression methods:
QOI, PNG, WebP, and JPEG-XL.  
QOI (Quite OK Image) is a simple and fast lossless image codec optimized for minimal computational overhead.  
PNG is a classic, widely adopted format that uses DEFLATE compression with Huffman coding.  
WebP and JPEG-XL are more modern formats that employ advanced prediction and entropy coding schemes, achieving higher compression efficiency at the cost of complexity.

We additionally evaluate SREC, a learned super-resolution-based compression method that relies on a large convolutional network.  
Finally, we include a variant of our method where the discrete lookup components (DLCs) are replaced with MLP layers of comparable parameter count, to examine how representational capacity affects compression efficiency.  
These learned baselines are considerably slower and not optimized for low-latency or energy-efficient operation.

\subsection{Training Details}  \label{appendix:hyperparams}

Both the ARM and UPS networks consist of two layers with 1024 lookup tables each.  
They are trained on 128,000 samples drawn from the EMNIST training set.  

The following summarizes the main hyperparameters used for training the ARM and UPS networks.  
During training, data augmentation was applied using random horizontal flips and random rotations of 0°, 90°, 180°, or 270°.

\begin{itemize}
    \item \textbf{Networks:} ARM and UPS, 2 layers of 1024 lookup tables each
    \item \textbf{Kernel size:} 5 for both ARM and UPS
    \item \textbf{Upsampling levels:} 2
    \item \textbf{Learning rate:} 0.01
    \item \textbf{Batch size:} 16
    \item \textbf{Training iterations:} 8,000
    \item \textbf{Connection temperature $\tau_{\text{connections}}$:} 1 $\to$ 0.0001, exponential decay by factor 10 every 2,000 iterations
    \item \textbf{Node temperature $\tau_{\text{node}}$:} 10 $\to$ 1, exponential decay by factor 10 every 2,000 iterations
    \item \textbf{Total samples seen:} 128,000
\end{itemize}

\section{Compression Difficulty of EMNIST Images}

\begin{figure*}[t]
    \centering
    \includegraphics[width=\linewidth]{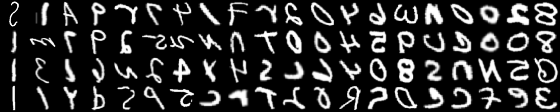}
    \caption{Examples from the EMNIST ByClass test set illustrating compression difficulty.  
    Left: images that are easy to encode, such as the letter \texttt{I}, which are straight and simple.  
    Right: images that are hard to encode, such as the number \texttt{8} and other characters with unusual or complex shapes.}
    \label{fig:difficulty}
\end{figure*}

Figure~\ref{fig:difficulty} illustrates examples from the EMNIST ByClass test set, highlighting the variation in compression difficulty across images.  
Some images are easier to encode due to their simple and regular structure; for example, the letter \texttt{I} is straight and uniform, which allows the model to predict per-pixel distributions with high confidence.  
In contrast, other images are more challenging to compress, such as the number \texttt{8} or characters with irregular or complex shapes.  
These images contain intricate details or unusual patterns, making accurate probability estimation harder and requiring more bits to represent losslessly.

\section{Energy and Latency Estimation Details}
\label{appendix:energy_estimation}

We estimate the energy usage of PNG and our method by combining direct measurements and literature-based extrapolations.  
For PNG, we measure the per-pixel energy consumption by timing the encoding and decoding operations and recording the corresponding increase in CPU power relative to idle.  
For our method, we do not have access to an FPGA implementation, so we base our estimates on reported measurements from related works.  
Specifically, we use published energy figures for small LUT-based neural networks of comparable size and assume similar energy characteristics for our model.

\subsection{PNG}
For PNG, we encode $1024\times1024$ grayscale images.  
We measure the average CPU power increase relative to idle and the time required to encode and decode a single pixel.  
Images are stored in RAM to avoid I/O overhead.  
Multiplying the measured power by the time per pixel yields the energy per pixel.  
Encoding takes 44.99\,ns and decoding 5.17\,ns.  
CPU power usage is 5.21\,J/s when idle, 12.38\,J/s during encoding, and 12.79\,J/s during decoding.  
The additional energy per pixel is 322.58\,nJ for encoding and 39.19\,nJ for decoding.  
This configuration favors PNG, since idle power is excluded and I/O latency is eliminated, allowing efficient bulk processing.

\subsection{Our Method}
DWN \citep{bacellar_differentiable_2025} measures 2.5\,nJ of energy consumption in a Xilinx Zynq Z-7045 FGPA for inferencing a 2.1k-LUT network.  
Our network is smaller (1024 LUTs), but we conservatively assume the same energy per inference.  
For level~0, the network is executed $28\times28$ times; for level~1, $2\times14\times14$ times; and for level~2, $2\times7\times7$ times.  
Normalizing by pixel yields
\[
\frac{28^2 + 2\cdot14^2 + 2\cdot7^2}{28^2} = 1.625,
\]
meaning the network runs on average 1.625 times per pixel.  
This results in approximately 4.06\,nJ per pixel for both encoding and decoding, which have roughly equal cost.

The asymmetric numerical system can be realized efficiently in hardware.  
tANS (table-based Asymmetric Numeral System) requires no DSPs, since it performs only table lookups for encoding and decoding.  
Even if we choose to use rANS (range-based ANS), the computation remains lightweight and can be implemented with only a few DSPs per symbol.

In rANS, each symbol $s$ with probability $P[s]$ and cumulative frequency $C[s]$ updates an internal integer state $x$.  
The normalization factor $F$ defines the range of valid states.  
During encoding, the current state is divided by the symbol probability to obtain the quotient and remainder:
\[
q = \left\lfloor \frac{x}{P[s]} \right\rfloor, \qquad
r = x \bmod P[s].
\]
The next state is then computed as
\[
x' = qF + C[s] + r.
\]

Decoding performs the inverse operations.  
From the encoded state $x'$, the decoder first computes
\[
m = x' \bmod F,
\]
then identifies the symbol $s$ such that $C[s] \le m < C[s+1]$.  
Finally, it reconstructs the previous state as
\[
x = P[s] \left\lfloor \frac{x'}{F} \right\rfloor + (x' \bmod F) - C[s].
\]

Both encoding and decoding involve only integer operations per symbol, mostly additions, multiplications, and modulo, with a single integer division per symbol.  
Assuming an energy cost of approximately 0.1\,nJ per operation \citep{rasoulinezhad2019pir, de2021fpga}, the total energy expenditure is negligible compared to LUT network inference.

Timing estimation depends on the FPGA layout.  
NeuraLUT \citep{andronic_neuralut_2024} reports a 3\,ns latency for a small 4000-LUT network.  
Upsampling could be performed in parallel, but even with sequential upsampling, both encoding and decoding remain below 5\,ns, making them faster than PNG encoding and comparable in decoding latency.

\end{document}